\documentclass[11pt,a4paper]{article}
\usepackage[hyphens]{url} % required to assure long URLs do not exceed the margins
\usepackage[hyperref]{ranlp2023}
\usepackage{times}
\usepackage{latexsym}

\usepackage[normalem]{ulem}
\useunder{\uline}{\ul}{}

\usepackage{amsmath,amssymb,amsfonts}
\usepackage{algorithmic}
\usepackage{graphicx}
\usepackage{multirow}
\usepackage[caption=false,font=footnotesize]{subfig}

\usepackage{microtype}

\aclfinalcopy % Uncomment this line for the final submission
\def\aclpaperid{150} %  Enter the acl Paper ID here

\title{From Fake to Hyperpartisan News Detection Using Domain Adaptation}

\author{Răzvan-Alexandru Smădu$^{1}$, Sebastian-Vasile Echim$^{1}$, Dumitru-Clementin Cercel$^{1}$, \\
\textbf{Iuliana Marin$^{2}$, Florin Pop$^{1,3}$}\\
$^{1}$University Politehnica of Bucharest, Faculty of Automatic Control and Computers\\
$^{2}$University Politehnica of Bucharest, Faculty of Engineering in Foreign Languages\\
$^{3}$National Institute for Research \& Development in Informatics - ICI Bucharest, Romania \\
\tt razvan.smadu@stud.acs.upb.ro, sebastian.echim@stud.aero.upb.ro\\
\tt \{dumitru.cercel,iuliana.marin,florin.pop\}@upb.ro\\
}

\date{}

\begin{document}
\maketitle
\begin{abstract}
Unsupervised Domain Adaptation (UDA) is a popular technique that aims to reduce the domain shift between two data distributions. It was successfully applied in computer vision and natural language processing. In the current work, we explore the effects of various unsupervised domain adaptation techniques between two text classification tasks: fake and hyperpartisan news detection. We investigate the knowledge transfer from fake to hyperpartisan news detection without involving target labels during training. Thus, we evaluate UDA, cluster alignment with a teacher, and cross-domain contrastive learning. Extensive experiments show that these techniques improve performance, while including data augmentation further enhances the results. In addition, we combine clustering and topic modeling algorithms with UDA, resulting in improved performances compared to the initial UDA setup.
\end{abstract}

\section{Introduction}

Fake news detection is a challenging task in which the goal is to detect whether the news content does not disseminate false information which may harm society. Recently, this problem has broad attention to the research community, especially with the rising interaction with social media platforms, which have become one of the primary sources of information for many individuals~\cite{shu2019fakenewsnet}. Detecting fake news is challenging for many of us, since some news can be written very convincingly, thus spreading misleading information without control~\cite{inproceedings-ISOT}. Therefore, new datasets (such as BuzzFeed-Webis Fake News (BuzzFeed)~\cite{buzzfeed} and ISOT~\cite{inproceedings-ISOT}) and novel detection techniques~\cite{koloski2022knowledge, mosallanezhad2022domain} have emerged in recent years.

%Another problem that has emerged, especially since the 2016 United States presidential election, is to identify whether the information spread by the news is in a political extreme.
Especially since the 2016 United States presidential election, a related task, namely hyperpartisan news detection, identifies whether the information spread by the news is in a political extreme~\cite{doi:10.1177/1461444820910416}. Hyperpartisan articles aim to expose information related to only one perspective, ignoring and, in some cases, even attacking the perspectives from other opposing sides~\cite{kiesel-etal-2019-semeval}. The consequences of this type of news range from misinformation in the media to an increase in the number of supporters of extreme ideologies~\cite{9034917}. 

Some works~\cite{buzzfeed,ross_rand_pennycook_2021} linked fake news with hyperpartisan news, since their goal is to spread as much as possible and influence people. This phenomenon is related to clickbait~\cite{potthast2016clickbait}, as the authors use different techniques to make the content more accessible and viral on the media~\cite{kiesel-etal-2019-semeval}. 

Recently, many architectures based on Bidirectional Encoder Representations from Transformers (BERT)~\cite{devlin2019bert} have been developed and fine-tuned on various natural language processing (NLP) tasks. The current work aims to evaluate unsupervised deep learning techniques on the fake news detection task and adapt them to the hyperpartisan news detection task. Specifically, we employ the Robustly optimized BERT 
pretraining approach
(RoBERTa)~\cite{liu2019roberta} and evaluate it in three domain adaptation scenarios: unsupervised domain adaptation (UDA)~\cite{ganin2015unsupervised}, cluster alignment with a teacher (CAT)~\cite{deng2019cluster}, and cross-domain contrastive learning (CDCL)~\cite{chen2020simple}. In addition, we analyze topic modeling and clustering algorithms to generate domain labels and perform UDA to learn about topic-aware features which are specific to fake and hyperpartisan news detection.  More precisely, we evaluate various clustering algorithms for generating domain labels, namely K-Means~\cite{KMeans}, K-Medoids \cite{kaufmann1987clustering}, Gaussian Mixture~\cite{fraley2002model}, and HDBSCAN~\cite{10.1007/978-3-642-37456-2_14}. Additionally, we explore four topic modeling algorithms: Latent Dirichlet Allocation (LDA)~\cite{blei2003latent}, Non-negative Matrix Factorization (NMF)~\cite{lee1999learning}, Latent Semantic Analysis (LSA)~\cite{deerwester1990indexing}, and probabilistic LSA (pLSA)~\cite{10.5555/2073796.2073829}.

% In this work, we aim to improve results on Hyperpartisan~\cite{kiesel-etal-2019-semeval} and Buzzfeed~\cite{buzzfeed} datasets for the hyperpartisan detection task.
Therefore, the main contributions of this work are as follows:
\begin{itemize}
    \item We evaluate the RoBERTa model on a domain adaptation from fake to hyperpartisan news detection by comparing three techniques, as well as several fine-tuning strategies.
    \item To our knowledge, we are the first to show that cross-domain contrastive learning proposed by \citet{wang2021crossdomain}, initially employed on computer vision, which performs better than other unsupervised learning techniques on an NLP task.
    \item We propose the cluster and topic-based UDA approaches, which obtain better results when compared with the original formulation for UDA.
    \item We perform extensive experiments to assess the effectiveness of each employed method under various hyperparameter configurations and data augmentation techniques based on the term frequency-inverse document frequency (TF-IDF) scores~\cite{salton1975vector}
    and the Generative Pre-trained Transformer 2 (GPT-2) model~\cite{radford2019language}.
\end{itemize}

\section{Related Work}
\label{sec:related_work}

% In this section, we present works concerning fake news, hyperpartisan detection, and domain adaptation techniques.

\subsection{Fake News Detection}
Machine learning techniques for detecting fake news include various feature-based methods, ranging from text to visual features~\cite{zhang2020overview}. For example, linguistic features~\cite{CHOUDHARY2021114171,pérezrosas2017automatic} capture aspects related to conveyed information, document organization, and vocabulary used in news. In contrast, style-based features~\cite{buzzfeed,zhou2020survey} are related to the writing style, such as redaction objectivity and deception~\cite{shu2017fake}. In recent years, Transformer-based models~\cite{vaswani2017attention} emerged in the fake news detection literature~\cite{jwa2019exbake,zhang2020bdann,kaliyar2021fakebert,szczepanski2021new}. Other techniques for detecting fake news use social aspects, such as the profiles of the users who spread the news on social media platforms~\cite{shu2017fake,onose2019hierarchical,zhou2020survey,sahoo2021multiple}. Techniques successfully employed for these scenarios rely on custom embeddings and linear classifiers~\cite{shu2019beyond}, classic supervised machine learning techniques~\cite{reis2019supervised}, and deep learning networks, such as recurrent~\cite{wu2018tracing} and graph neural networks~\cite{monti2019fake, hamid2020fake, paraschiv2021graph}.

\subsection{Hyperpartisan News Detection}
%As a binary classification task, hyperpartisan detection from news articles was introduced at the SemEval-2019 Task 4 challenge~\cite{kiesel-etal-2019-semeval}. 
% analyzed
Task 4 of SemEval-2019~\cite{kiesel-etal-2019-semeval} introduced hyperpartisan detection from news articles as a binary classification task. The organizers created two balanced datasets by crawling data from various online publishers. Participants were asked to detect whether the news articles were hyperpartisan or mainstream. The winning team~\cite{jiang-etal-2019-team} of the shared task proposed an architecture based on multiple pre-trained ELMo embeddings~\cite{peters2019knowledge} averaged in the embedding space, followed by convolutional layers~\cite{zhang2015sensitivity} and batch normalization~\cite{ioffe2015batch}. They achieved 84.04\% accuracy on the training set and 82.16\% accuracy on the test set, suggesting the challenging setting. Other works for the SemEval-2019 Task 4 were based on lexical and semantic handcrafted features via Universal Sentence Encoder~\cite{cer2018universal} or BERT, and a linear classifier~\cite{srivastava2019vernon,hanawa-etal-2019-sally}. Furthermore, \citet{buzzfeed} showed that hyperpartisan news detection could be analyzed using fake news approaches. They argued that the writing style for hyperpartisan news is similar to fake news, despite their political orientation.

\subsection{Unsupervised Domain Adaptation}
The core objective of unsupervised domain adaptation is to enforce a feature representation invariant to the domain of the examples with the same labels. One of the most effective techniques is the work of \citet{ganin2015unsupervised}, which treated the problem as a minimax optimization.
%Other unsupervised approaches were also developed~\cite{wang2018eann,deng2019cluster}, some of which were investigated in the fake news detection task. For example,
\citet{wang2018eann} utilized domain adaptation techniques via adversarial training for fake news detection by employing an event discriminator to learn event-invariant features in a multi-modal setting. \citet{deng2019cluster} relied on the similarity in the feature space by enforcing a clustered structure among similar features. In this case, the training procedure optimizes clustering loss alongside the domain adaptation loss. For the target dataset, a teacher model consisting of an ensemble of students generates pseudo-labels (i.e., estimates of the true labels). Also, contrastive learning~\cite{chen2020simple} was used to achieve unsupervised domain adaptation. It aims to have closer representations of the examples from the same class, while representations from different classes should stay far apart. In addition, \citet{wang2021crossdomain} proposed the cross-domain contrastive loss to minimize the $l_2$-norm distance between features from the same category, and employed \mbox{K-Means} to compute pseudo-labels.

\begin{figure*}[!ht]
    \centering
    \includegraphics[width=0.85\textwidth]{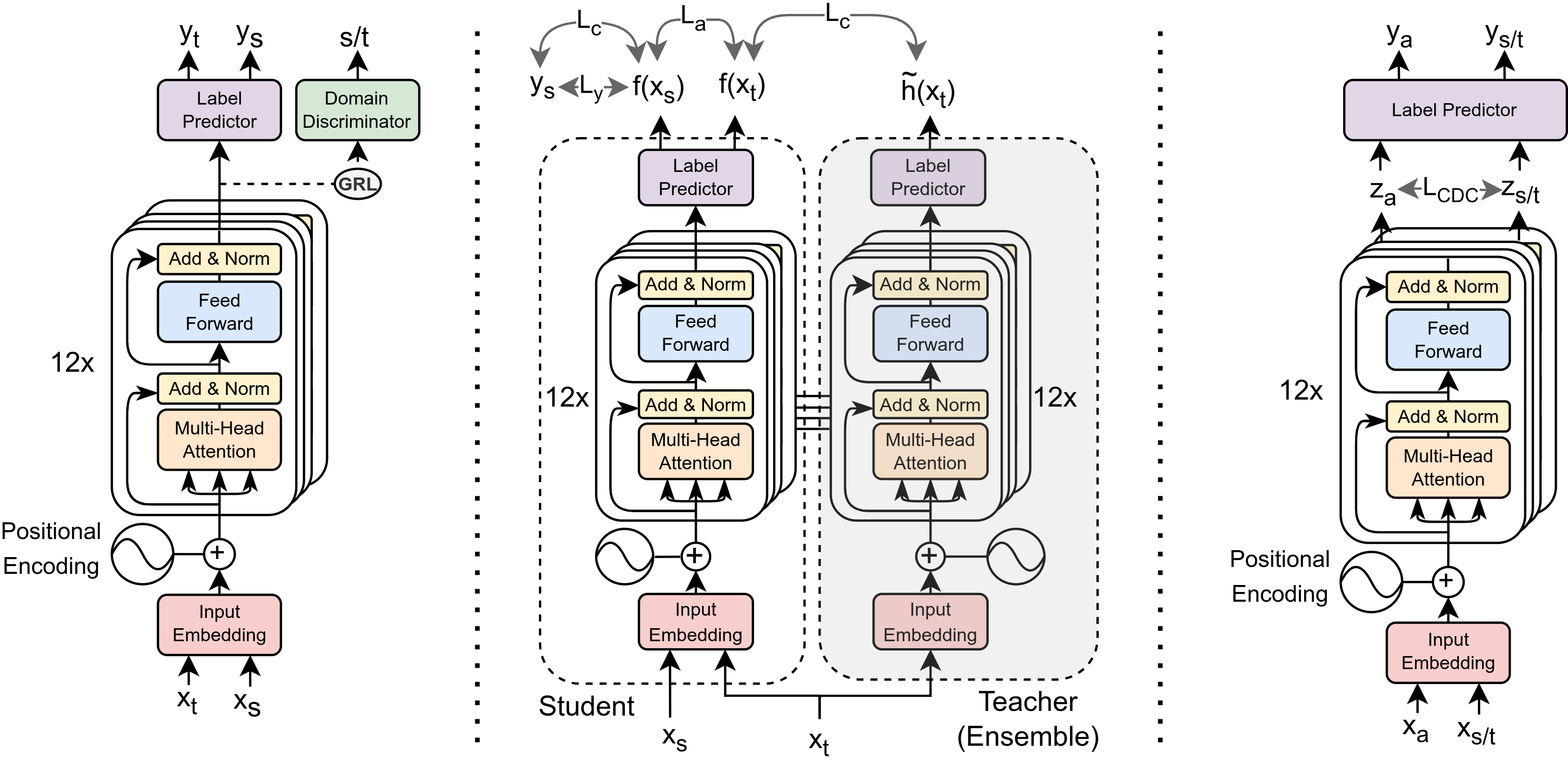}
    \caption{
    %The RoBERTa model used in the following settings: UDA (left), CAT (center), and CDCL (right).
    (Left) The RoBERTa model in the UDA setting includes a label predictor and a domain discriminator. (Center) In the CAT method, the student and teacher use the RoBERTa model. (Right) In the CDCL setting, the contrastive loss is applied between the RoBERTa features of an anchor and the source (s) / target (t) example.}
    \label{fig:method}
\end{figure*}

\section{Method}
\label{sec:approach}

% In this section, we describe the models and the methods used to perform domain adaptation between a source and a target domain. 

\subsection{Base Model}

In our current work, we utilize the pre-trained RoBERTa language 
model, which shares the same architectural design as BERT, the only difference being the pre-training objectives. The RoBERTa architecture stacks multiple Transformer encoders, each based on the multi-head self-attention mechanism~\cite{vaswani2017attention}. On top of the RoBERTa model, we add a label predictor containing fully connected layers. %In contrast to BERT,
RoBERTA uses the Byte-Pair Encoding (BPE) tokenizer~\cite{sennrich2015neural}. In what follows, we present the settings in which RoBERTa is employed in our work (see Figure~\ref{fig:method}).

\subsection{Unsupervised Domain Adaptation}

Given two datasets $D_s=\{(x^i_s,y^i_s)\}_{i=1}^{N_s}$ and $D_t=\{x^i_t\}_{i=1}^{N_t}$ from different domains, the UDA setting reduces the shift between them~\cite{ganin2015unsupervised,ganin2016domainadversarial}. This approach comprises a feature encoder $G_f$, a label predictor $G_y$, and a domain discriminator $G_d$. The feature encoder maps the input space into a latent space. Then, the label predictor computes the labels of the underlying examples. Simultaneously, the domain classifier uses the latent space to predict the domain of the features (i.e., the source or target domain).

To obtain domain-invariant features, the optimization is two-fold. First, we minimize the prediction loss concerning $G_f$'s parameters $\theta_f$ and $G_y$'s parameters $\theta_y$. Second, we maximize the domain classification loss until $G_d$ cannot distinguish the domains of the features. Formally, the loss function $L$ 
(see Eq. \ref{eq:loss}) depends on the prediction loss $L_y$ between $G_y$'s outputs and source labels, and the domain adaptation loss $L_d$ between $G_d$'s outputs and domains $d^i$ (i.e., hyperpartisan and fake news). The trade-off between $L_y$ and $L_d$ is controlled by $\lambda$. Note that we omitted the model's parameters for clarity. 
\begin{equation}
\begin{split}
    \label{eq:loss}
    L &= \sum_{i=1}^{N_s} L_y(G_y(G_f(x^i_s)), y^i_s) \\
      &- \lambda \sum_{i=1}^{N} L_d(G_d(G_f(x^i)), d^i)
\end{split}
\end{equation}

The optimization problem associated with this formulation is described below:
\begin{equation}
    \hat{\theta}_f, \hat{\theta}_y = \arg\min_{\theta_f,\theta_y} L(\theta_f, \theta_y, \hat{\theta}_d)
\end{equation}
\begin{equation}
    \hat{\theta}_d = \arg\max_{\theta_d} L(\hat{\theta}_f, \hat{\theta}_y, \theta_d)
\end{equation}
where the parameters with hat are fixed during the optimization step. This problem can be solved with an implementation trick, namely gradient reversal layer (GRL)~\cite{ganin2015unsupervised}, which acts as the identity function during feed-forward and negates the gradients during back-propagation. The GRL layer is inserted between the feature encoder and the domain discriminator.

In our setting, we use the RoBERTa's encoders for feature extraction and fully connected layers for both the label predictor and domain discriminator.

\subsection{Cluster Alignment with a Teacher}

As an extension to UDA, \citet{deng2019cluster} exploited the class-conditional structure of the feature space by cluster alignment in the teacher-student paradigm. A teacher model trained on the labeled  source examples estimates pseudo-labels for the unlabeled target dataset.
%to increase the separability among the target labels.
To reduce the error amplification caused by label estimation, the teacher model is built as an ensemble of previous student classifiers. In addition, a student classifier minimizes the prediction loss $L_y$ on the source examples in the supervised setting. The optimization involves minimizing both the prediction loss $L_y$ and the sum of clustering losses $L_c$ (i.e., for both the source and the target domains) and the cluster-base alignment loss $L_a$:
\begin{equation}
\label{eq:CAT_loss}
    L = L_y + \alpha (L_c + L_a)
\end{equation}
where the hyperparameter $\alpha$ controls the trade-off between the supervised and semi-supervised losses.

Considering  the labeled samples $X_s = \{x_s^i,y_s^i\}_{i=1}^{N_s}$, the unlabeled samples $X_t = \{x_t^i\}_{i=1}^{N_t}$, the feature extractor $f(\cdot)$, and the distance metric $d$ between features, the total clustering loss is:
\begin{equation}
    L_c(X_s, X_t) = L_c(X_s) + L_c(X_t)
\end{equation}
where $L_c$ is as follows for each $X_*$:

\begin{equation}
\begin{split}
    & L_c(X_*) = \frac{1}{\left| X_* \right|^2} \sum_{i=1}^{\left| X_* \right|} \sum_{j=1}^{\left| X_* \right|} [\delta_{ij} d(f(x^i), f(x^j)) \\ 
           &+ (1-\delta_{ij}) \max (0, m - d(f(x^i), f(x^j))) ]
\end{split}
\end{equation}

The intuition is to enforce class-conditional structure at the feature representation by grouping the classes into clusters, i.e., by minimizing the distance between features $x^i$ and $x^j$ that have the same label when the indicator function $\delta_{ij}=1$, whereas pushing different clusters away from at least a margin $m$ by maximizing the feature distance  when $\delta_{ij}=0$. The classifier trained on the source features may not be able to differentiate between the same class from different domains, and therefore, an alignment loss $L_a$ is imposed between the domains as follows:
 
\begin{equation}
    L_a(X_s, X_t) = \frac{1}{K} \sum_{k=1}^K \left|\left| \lambda_{s,k} - \lambda_{t,k} \right|\right|_2^2
\end{equation}

In this case, given the number $K$ of classes to be predicted, and the samples $X_{*,k}$ from either source or target whose labels are equal to $k$, the cluster centroids $\lambda_{*,k}$ are computed using:

\begin{equation}
    \lambda_{*,k} = \frac{1}{\left| X_{*,k} \right|} \sum_{x^i_{*} \in X_{*,k}} f(x^i_{*})
\end{equation}

The loss $L_a$  tries to match the source and target statistics by aligning the clusters for each class $k$ in the feature space. Additionally, the performance can be further improved by aligning the marginal distributions, i.e., adding a confidence threshold that ignores the data points likely to be included in the wrong class.

\subsection{Cross-Domain Contrastive Learning}

Self-supervised contrastive learning~\cite{chen2020simple} aims to learn representations such that, given a pair of examples, closely related examples should behave similarly, while dissimilar examples should stay far apart from each other. This can be achieved by employing various techniques such as data augmentation and custom losses (e.g., NT-Xent~\cite{chen2020simple}, InfoNCE~\cite{oord2019representation}). Since there is no clear way to construct positive and negative pairs in an unsupervised domain adaptation framework, \citet{wang2021crossdomain} argued that samples from the same category should be similar. In contrast, samples from different categories should have other feature representations, regardless of the domain from which they come. Based on this hypothesis, they proposed the cross-domain contrastive (CDC) loss to reduce the domain shift between source and target labels. 
We assume $z^a_t$ and $z^p_s$ are the $l_2$-normalized features for the anchor sample from the target domain $x^a_t$ and the positive sample from the source domain $x^p_s$, respectively. In this case, the loss function is described by: 

\begin{equation}
\label{eq:cdc_loss}
L^{t,a}_{CDC} = -\frac{1}{\left| P_s(\hat{y}^a_t) \right|} \sum_{p \in P_s(\hat{y}^a_t)} log \frac{ \exp({{z^a_t} \cdot z^p_s / \tau})}{\sum\limits_{j \in I_s} \exp({{z^a_t} \cdot z^j_s / \tau})}
\end{equation}
where $P_s(\hat{y}^a_t)$ denotes the set of positive samples from the source domain having the same label as the anchor point, and $I_s$ is the set of all source samples from the mini-batch. Similar to Eq.~\ref{eq:cdc_loss}, we compute  $L^{s,a}_{CDC}$, for which we consider the positive samples from the target domain instead. The CDC loss with alignment at the feature level is\footnote{Note that we included the normalization terms compared to the original formulation.}:

\begin{equation}
    L_{CDC} = \frac{1}{N_s} \sum_{a=1}^{N_s} L^{s,a}_{CDC} + \frac{1}{N_t} \sum_{a=1}^{N_t} L^{t,a}_{CDC}
\end{equation}

The objective function is given by the sum of the prediction loss $L_y$ and the loss $L_{CDC}$ scaled by $\gamma$: 
\begin{equation}
    L = L_y + \gamma L_{CDC}
\end{equation}

We generate pseudo-labels using the K-Means algorithm since we require them when creating positive pairs. We initialize K-Means with the centroids of the source domain and predict on the target domain. The pseudo-labels are chosen to minimize the similarity distance between the feature representation and the centroid. K-Means is performed at the beginning of each epoch.

\subsection{Cluster and Topic-Based Unsupervised Domain Adaptation}
% Something about clustering.

We propose an addition to the UDA approach, considering the supervised setting (i.e., we have access to the labeled source dataset). First, we represent the input text using TF-IDF or a pre-trained RoBERTa model. We employ a clustering/topic modeling algorithm in this feature space to identify $k$ clusters or topics, which will be assigned as domain labels. For clustering, we employ four algorithms, namely K-Means, K-Medoids, Gaussian Mixture, and HDBSCAN. Also, we use four topic modeling algorithms, namely LDA, NMF, LSA, and pLSA.
The motivation is to compact the latent representation, given estimates of latent domains under a topic model (i.e., a dataset split). During training, it is minimized the loss given by Eq.~\ref{eq:loss} while using the proposed domain labels. For the target examples, we do not include labels during training. We choose the number of clusters using the elbow method\footnote{\url{https://www.scikit-yb.org/en/latest/api/cluster/elbow.html}}. After training on each pair of domain labels, the best-performing model is selected for the inference stage.

\section{Experimental Setup}
\label{sec:experiments}

\subsection{Datasets}

%We evaluate the previously presented  approaches and perform experiments on three datasets related to fake news (i.e., ISOT Fake News, and BuzzFeed-Webis Fake News) and hyperpartisan detection (i.e., BuzzFeed-Webis Fake News, and Hyperpartisan).
We perform experiments on three datasets related to fake (i.e., ISOT and BuzzFeed) and hyperpartisan (i.e., BuzzFeed and {Hyperpartisan}~\cite{kiesel-etal-2019-semeval}) news detection.

The ISOT fake news dataset contains news articles collected from reuters.com, and other websites, which were validated by Politifact\footnote{An organization that checks the veracity of the news.}. The dataset comprises 44,898 articles, of which 21,417 contain truthful information, and 23,481 are fake news. All collected articles are related to politics and have at least 200 characters.

The BuzzFeed dataset contains 1,627 articles in three categories: mainstream, left-wing, and right-wing. The mainstream and hyperpartisan data are evenly distributed, and the length of the articles ranges between 400 and 800 words. This dataset is annotated for both fake and hyperpartisan news detection.

The Hyperpartisan dataset which contains hyperpartisan news was released under the SemEval-2019 Task 4 shared task~\cite{kiesel-etal-2019-semeval}. The dataset was crawled from news publishers listed by \mbox{BuzzFeed}\footnote{\url{https://github.com/BuzzFeedNews/2017-08-partisan-sites-and-facebook-pages}} and Media Bias Fact Check\footnote{\url{https://mediabiasfactcheck.com}}. From these sources, 754,000 news articles were extracted and semi-automated labeled using distant supervision \cite{mintz-etal-2009-distant} at the publisher level, provided in the HTML format. It was split into 600,000 articles for training, 150,000  articles for validation, and 4,000 articles for testing. Half of the dataset consists of non-hyperpartisan articles, and the other half is split equally among left-wing and right-wing articles. Since the authors also released a smaller version of the dataset (645 examples for training and 628 examples for testing), in what follows, we will refer to the larger dataset as Hyperpartisan-L and the smaller dataset as Hyperpartisan-S.

\subsection{Data Preprocessing}

We perform data cleaning on all three corpora, ignoring non-ASCII characters and removing HTML-specific symbols and constructions that do not provide any information about the actual content, such as multiple chains of dots in a line. 
%We utilized the BPE for tokenization which was set to output a maximum of 128 tokens per text sample.
BPE was utilized for tokenization, setting to output a maximum of 128 tokens per text sample.
 
Since the ISOT and BuzzFeed datasets are not provided with separate splits for validation and testing, we use the following split: 70\% for training, 10\% for validation, and 20\% for testing. In addition, due to limited computational resources and the large size of the Hyperpartisan dataset, we select a random 5\% of the data from the training set (i.e., 30,000 examples) and  5\% of the data for the validation set (i.e., 7,500 examples). Also, we use the entire Hyperpartisan test set since it contains only 4,000 examples.

\subsection{Hyperparameters}

We utilize the pre-trained RoBERTa base version (123M parameters), which consists of a stack of 12 Transformer blocks. For all experiments, the Adam optimizer~\cite{kingma2017adam} with a linear scheduler is used with a warm-up (it is set with 5\% of the gradient steps) for the learning rate. The learning rate varies among experiments, between $1e-4$ and $1e-5$. We employ a dropout set between 0.1 and 0.5. We also set the optimizer's weight decay parameter to $1e-3$, and clip the gradients between -1 and 1 to increase training stability and reduce overfitting.

\section{Results}
\label{sec:results}

There were conducted multiple experiments to evaluate the impact of using various fine-tuned models for RoBERTa. We also investigate the effects of fine-tuning the RoBERTa model on the downstream task. Then, we analyze the impact of using a data augmentation technique~\cite{xie2020unsupervised} based on the TF-IDF scores. In Appendix~\ref{app:gpt2_results}, we present the results of the GPT-2 data augmentation. Finally, we use clustering and topic modeling algorithms to extract clusters and topics from the training set and perform domain adaptation. We present the results in terms of accuracy (Acc) and F1-score (F1).

\subsection{Baselines}
We start with the most straightforward approach for training a neural network. That is, we take a pre-trained model on similar tasks and transfer some of the acquired knowledge to the downstream task via fine-tuning. The baseline model consists of the RoBERTa model followed by a stack of fully connected layers. We employ two fully connected layers, with 256 hidden units and two output neurons. The models are trained for 3 epochs, with a learning rate of $1e-4$ and batch size between 32 and 64.
%Since we deal with a binary classification task, w

First, we evaluate the model on all four datasets for baseline results. Table~\ref{tab:baseline-results-1} presents the final results obtained during experiments. We observe that ISOT achieves the highest scores, followed by BuzzFeed and Hyperpartisan-S. We note that humans annotated these datasets, whereas the Hyperpartisan-L dataset was annotated with a semi-supervised approach. 
% In addition, we observe that RoBERTa does not obtain very high precision because of a high false positives rate.

By comparing three fine-tuning methods (see Table~\ref{tab:baseline-results-2}), we observe that freezing the model's encoders yields poor performance. This increases the number of false positives and decreases the number of true negatives because of the domain shift between the datasets and training with fewer parameters. On the other hand, fine-tuning improves the results since the model's parameters are adapted to the new domain.

\begin{table}[!t]
\centering
\caption{\label{tab:baseline-results-1} Results obtained after fine-tuning and evaluating RoBERTa on each dataset.}
\resizebox{0.7\columnwidth}{!}{
\begin{tabular}{l|c|c}
\hline
\textbf{Dataset} & \textbf{Acc(\%)} & \textbf{F1(\%)} \\ \hline
BuzzFeed         & 96.9             & 96.7             \\
ISOT            & 99.8             &  99.7             \\
Hyperpartisan-S        & 83.7             & 83.0             \\
Hyperpartisan-L        & 62.1             & 69.0             \\
% \textbf{Dataset} & \textbf{Acc(\%)} & \textbf{P(\%)} & \textbf{R(\%)} & \textbf{F1(\%)} \\ \hline
% BuzzFeed         & 96.9              & 95.5            & 98.0            & 96.7             \\
% ISOT             & 99.8              & 99.7            & 99.7            & 99.7             \\
% Hyperpartisan-S        & 83.7              & 83.3            & 82.7            & 83.0             \\
% Hyperpartisan-L        & 62.1              & 59.2            & 82.8            & 69.0             \\
\hline
\end{tabular}}
\end{table}

\begin{table}[!t]
\centering
\caption{\label{tab:baseline-results-2} Results for different fine-tuning strategies on the Hyperpartisan-L dataset.}
\resizebox{\columnwidth}{!}{
\begin{tabular}{l|c|c}
\hline
\textbf{Model}             & \textbf{Acc(\%)} & \textbf{F1(\%)} \\ \hline
RoBERTa                    & 62.1                          & 69.0             \\
RoBERTa frozen             & 53.7                          & 65.4             \\
RoBERTa fine-tuned first on BuzzFeed & 62.3                & 68.0             \\
RoBERTa fine-tuned first on ISOT     & \textbf{63.0}                   & \textbf{70.0}             \\
% RoBERTa           & 62.1              & \textbf{59.2}            & 82.8            & 69.0             \\
% RoBERTa frozen             & 53.7              & 52.2            & \textbf{87.7}            & 65.4             \\
% RoBERTa fine-tuned first on BuzzFeed & 62.3              & 59.0            & 80.4            & 68.0             \\
% RoBERTa fine-tuned first on ISOT     & \textbf{63.0}              & 58.8            & 86.4            & \textbf{70.0}             \\
\hline
\end{tabular}}
\end{table}

\subsection{Results for UDA}

We consider the encoders from the RoBERTa model as feature generators. We also use a stack of fully connected layers, with 256 hidden neurons and two outputs for both the label predictor and the domain discriminator. The domain discriminator is linked to the output of the RoBERTa encoder via a gradient reversal layer. We tested three values for $\lambda \in \{0.1, 1, 5\}$.

Furthermore, we perform larger-to-smaller and smaller-to-larger dataset adaptations between Hyperpartisan-L and BuzzFeed. The model is trained for 3 epochs (i.e., the steps required to pass through all examples from the larger dataset). The batch size is set to 64, from which half are labeled and the other half are unlabeled examples. The results are shown in Table \ref{tab:UDA_results}.
% Performing the larger-to-smaller adaptation increases the accuracy while reducing the precision. In the case of performing smaller-to-larger adaptation, the model is more prone to introducing false negatives or positives, resulting in very high precision and very low recall, or vice-versa.
We observe that if $\lambda$ is set too large, the model does not learn the data distribution but predicts only one class. Conversely, UDA performs better when $\lambda=0.1$, achieving higher accuracy on the Hyperpartisan-L target dataset. This adaptation may have helped because of the inherent similarities between domains and improved performance on out-of-distribution points.

Moreover, we employ different ways of linking the GRL layer with the RoBERTa encoders. Since the RoBERTa-base model uses 12 encoders, we utilized the 4th, 6th, and 10th, besides the previous experiments. While the encoder returns a feature representation for each element in the sequence, we take the representation of the \texttt{[CLS]} token. Table~\ref{tab:GRL_results} shows the results. The 12th layer performs best, while similar performances are achieved using the 4th or 6th layer. The results are supported by the fact that more layers for the encoder mean more representational power for the feature encoder that needs to be adapted among domains. 

\begin{table}[!t]
\centering
\caption{\label{tab:UDA_results} Unsupervised domain adaptation between Hyperpartisan-L and BuzzFeed datasets.}
\resizebox{\columnwidth}{!}{
\begin{tabular}{l|l|l|c|c|c|c}
\hline
\multicolumn{1}{c|}{\multirow{2}{*}{\textbf{$\lambda$}}} & \multicolumn{1}{c|}{\multirow{2}{*}{\textbf{Source}}} & \multicolumn{1}{c|}{\multirow{2}{*}{\textbf{Target}}} & \multicolumn{2}{c|}{\textbf{Source}}                                                                                       & \multicolumn{2}{c}{\textbf{Target}}                                                                                                                  \\ \cline{4-7}
\multicolumn{1}{c|}{}                                                & \multicolumn{1}{c|}{}                                 & \multicolumn{1}{c|}{}                                 & \multicolumn{1}{c|}{\textbf{Acc(\%)}} & \multicolumn{1}{c|}{\textbf{F1(\%)}} & \multicolumn{1}{c|}{\textbf{Acc(\%)}} & \multicolumn{1}{c}{\textbf{F1(\%)}} \\ \hline
0.1 & Hyperpartisan-L & BuzzFeed & \textbf{61.5} & 67.7 & \textbf{85.4} & \textbf{86.4} \\
1 & Hyperpartisan-L & BuzzFeed & 58.1 & \textbf{68.4} & 60.8 & 38.2 \\
5 & Hyperpartisan-L & BuzzFeed & 50.0  & 2.5 & 54.0 & 3.8 \\ \hline
0.1 & BuzzFeed & Hyperpartisan-L & 95.3 & 94.9 & \textbf{64.3} & 62.7 \\
1 & BuzzFeed & Hyperpartisan-L & \textbf{96.5} & \textbf{96.6} & 50.0 & \textbf{66.5} \\
5 & BuzzFeed & Hyperpartisan-L & 51.5 & 7.1 & 50.8 & 7.7 \\ \hline
0.1 & BuzzFeed & Hyperpartisan-L & 94.4 & 94.5 & 56.7 & 64.1 \\
% \multicolumn{1}{|c|}{}                                                & \multicolumn{1}{c|}{}                                 & \multicolumn{1}{c|}{}                                 & \multicolumn{1}{c|}{\textbf{Acc(\%)}} & \multicolumn{1}{c|}{\textbf{P(\%)}} & \multicolumn{1}{c|}{\textbf{R(\%)}} & \multicolumn{1}{c|}{\textbf{F1(\%)}} & \multicolumn{1}{c|}{\textbf{Acc(\%)}} & \multicolumn{1}{c|}{\textbf{P(\%)}} & \multicolumn{1}{c|}{\textbf{R(\%)}} & \multicolumn{1}{c|}{\textbf{F1(\%)}} \\ \hline
% 0.1 & Hyperpartisan-L & BuzzFeed & \textbf{61.5} & \textbf{58.3} & 80.9 & 67.7 & \textbf{85.4} & 80.6 & \textbf{93.1} & \textbf{86.4} \\
% 1 & Hyperpartisan-L & BuzzFeed & 58.1 & 54.9 & \textbf{90.7} & \textbf{68.4} & 60.8 & 72.2 & 26.0 & 38.2 \\
% 5 & Hyperpartisan-L & BuzzFeed & 50.0 & 50.9 & 1.3 & 2.5 & 54.0 & \textbf{100} & 1.9 & 3.8 \\ \hline
% 0.1 & BuzzFeed & Hyperpartisan-L & 95.3 & 96.5 & 93.3 & 94.9 & \textbf{64.3} & \textbf{65.6} & 60.1 & 62.7 \\
% 1 & BuzzFeed & Hyperpartisan-L & \textbf{96.5} & 95.2 & \textbf{98.1} & \textbf{96.6} & 50.0 & 50.0 & \textbf{99.2} & \textbf{66.5} \\
% 5 & BuzzFeed & Hyperpartisan-L & 51.5 & \textbf{100} & 3.7 & 7.1 & 50.8 & 61.9 & 4.1 & 7.7 \\ \hline
% 0.1 & BuzzFeed & Hyperpartisan-L & 94.4 & 93.1 & 94.8 & 94.5 & 56.7 & 54.7 & 77.4 & 64.1 \\
\hline
\end{tabular}}
\end{table}
 % On the last line separated from the rest of the results, the model was fine-tuned from the pre-trained RoBERTa on ISOT.

\begin{table}[!t]
\centering
\caption{\label{tab:GRL_results} Various linking positions of the GRL layer to the encoders of RoBERTa, on BuzzFeed (source) to Hyperpartisan-L (target) adaptation.}
\resizebox{0.8\columnwidth}{!}{
\begin{tabular}{l|c|c|c|c}
\hline
\multicolumn{1}{c|}{\multirow{2}{*}{\textbf{\begin{tabular}[c|]{@{}c@{}}GRL\\ pos.\end{tabular}}}} & \multicolumn{2}{c|}{\textbf{Source}}                                                                                              & \multicolumn{2}{c}{\textbf{Target}}                                                                                              \\ \cline{2-5}
\multicolumn{1}{c|}{}                                                                             & \multicolumn{1}{c|}{\textbf{Acc(\%)}} & \multicolumn{1}{c|}{\textbf{F1(\%)}} & \multicolumn{1}{c|}{\textbf{Acc(\%)}} & \multicolumn{1}{c}{\textbf{F1(\%)}} \\ \hline
4 & \textbf{95.9} & \textbf{95.2} & 62.1 & 61.7 \\
6 & 95.0 & 94.4 & 62.1 & \textbf{67.1} \\
10 & 91.3 & 89.1 & 60.9 & 64.1 \\
12 & 95.3 & 94.9 & \textbf{64.3} & 62.7 \\

% \multicolumn{1}{|c|}{\multirow{2}{*}{\textbf{\begin{tabular}[c|]{@{}c@{}}GRL\\ pos.\end{tabular}}}} & \multicolumn{4}{c|}{\textbf{Source}}                                                                                              & \multicolumn{4}{c|}{\textbf{Target}}                                                                                              \\ \cline{2-9}
% \multicolumn{1}{|c|}{}                                                                             & \multicolumn{1}{c|}{\textbf{Acc(\%)}} & \multicolumn{1}{c|}{\textbf{P(\%)}} & \textbf{R(\%)} & \multicolumn{1}{c|}{\textbf{F1(\%)}} & \multicolumn{1}{c|}{\textbf{Acc(\%)}} & \multicolumn{1}{c|}{\textbf{P(\%)}} & \textbf{R(\%)} & \multicolumn{1}{c|}{\textbf{F1(\%)}} \\ \hline
% 4 & \textbf{95.9} & \textbf{97.7} & 92.8 & \textbf{95.2} & 62.1 & 62.3 & 61.0 & 61.7 \\
% 6 & 95.0 & 91.8 & \textbf{97.1} & 94.4 & 62.1 & 59.3 & \textbf{77.3} & \textbf{67.1} \\
% 10 & 91.3 & 96.6 & 82.7 & 89.1 & 60.9 & 59.3 & 69.7 & 64.1 \\
% 12 & 95.3 & 96.5 & 93.3 & 94.9 & \textbf{64.3} & \textbf{65.6} & 60.1 & 62.7 \\

\hline
\end{tabular}}
\end{table}

\subsection{Results for CAT}

In addition to the previous experimental setup, we set the parameter $\alpha \in \{0.1, 1\}$ for the clustering loss in the CAT configuration. We also consider a lower learning rate (i.e., $1e-5$) to improve convergence. 
% The teacher model is an ensemble of previous student models. Since we use a teacher model for the pseudo-labels,
We consider an epoch is a complete pass through the smaller dataset to update the pseudo-labels for the entire target domain using the teacher model. As such, we trained the models for 10-30 epochs. We set the margin $m=2$, the ensemble size to 3, and the ensemble accumulation to 0.8.

We performed domain adaptation from BuzzFeed to Hyperpartisan-L.
The results are shown in Table \ref{tab:cat_results}. The model obtains over 90\% accuracy on the source domain and is bounded by 66.4\% on the target domain. This approach generally achieves a smaller accuracy than previous techniques, the best score being when $\lambda = \alpha = 0.1$. Also, we can observe that the difference between $\lambda$ and $\alpha$ affects the performances. Analyzing the model predictions, we notice that using smaller values for $\lambda$ and $\alpha$ yields a high number of false positives, while larger values increase the number of false negatives. Using $\lambda=1$ and $\alpha=0.1$ resulted in a biased model towards mainstream examples.

\begin{table}[!t]
\centering
\caption{\label{tab:cat_results} Results for the CAT framework on BuzzFeed (source) to Hyperpartisan-L (target) adaptation.}
\resizebox{0.8\columnwidth}{!}{
\begin{tabular}{l|l|c|c|c|c}
\hline
\multicolumn{1}{c|}{\multirow{2}{*}{\textbf{$\lambda$}}} & \multicolumn{1}{c|}{\multirow{2}{*}{\textbf{$\alpha$}}} & \multicolumn{2}{c|}{\textbf{Source}}                                                                                          & \multicolumn{2}{c}{\textbf{Target}} \\ \cline{3-6}
\multicolumn{1}{c|}{}                                 & \multicolumn{1}{c|}{}                                & \multicolumn{1}{c|}{\textbf{Acc(\%)}} & \multicolumn{1}{c|}{\textbf{F1(\%)}} & \multicolumn{1}{c|}{\textbf{Acc(\%)}} & \multicolumn{1}{c}{\textbf{F1(\%)}} \\ \hline
1 & 1 & 92.5 & 91.2 & 51.3 & \textbf{66.4} \\
1 & 0.1 & 94.7 & 93.8 & 57.9 & 62.6 \\
0.1 & 0.1 & 95.9 & 95.7 & \textbf{59.9} & 61.5 \\
0.1 & 0 & \textbf{96.5} & \textbf{96.4} & 58.7 & 64.3 \\
0 & 0.1 & 95.6 & 95.4 & 59.8 & 64.1 \\
0 & 0 & 93.7 & 92.7 & 58.9 & 62.5 \\

% \multicolumn{1}{|c|}{\multirow{2}{*}{\textbf{$\lambda$}}} & \multicolumn{1}{c|}{\multirow{2}{*}{\textbf{$\alpha$}}} & \multicolumn{4}{c|}{\textbf{Source}}                                                                                          & \multicolumn{4}{c|}{\textbf{Target}} \\ \cline{3-10}
% \multicolumn{1}{|c|}{}                                 & \multicolumn{1}{c|}{}                                & \multicolumn{1}{c|}{\textbf{Acc(\%)}} & \multicolumn{1}{c|}{\textbf{P(\%)}} & \textbf{R(\%)} & \multicolumn{1}{c|}{\textbf{F1(\%)}} & \multicolumn{1}{c|}{\textbf{Acc(\%)}} & \multicolumn{1}{c|}{\textbf{P(\%)}} & \textbf{R(\%)} & \multicolumn{1}{c|}{\textbf{F1(\%)}} \\ \hline
% 1 & 1 & 92.5 & 92.5 & 89.9 & 91.2 & 51.3 & 50.7 & \textbf{96.5} & \textbf{66.4} \\
% 1 & 0.1 & 94.7 & 94.8 & 92.8 & 93.8 & 57.9 & 56.3 & 70.4 & 62.6 \\
% 0.1 & 0.1 & 95.9 & \textbf{96.7} & 94.8 & 95.7 & \textbf{59.9} & \textbf{59.0} & 64.2 & 61.5 \\
% 0.1 & 0 & \textbf{96.5} & 95.5 & \textbf{97.4} & \textbf{96.4} & 58.7 & 56.6 & 74.6 & 64.3 \\
% 0 & 0.1 & 95.6 & \textbf{96.7} & 94.2 & 95.4 & 59.8 & 57.8 & 72.0 & 64.1 \\
% 0 & 0 & 93.7 & 93.4 & 92.0 & 92.7 & 58.9 & 57.5 & 68.5 & 62.5 \\

\hline
\end{tabular}}
\end{table}

\subsection{Results for CDCL}

% we generate pseudo-labels using K-Means on the feature space. Thus, before the epoch begins, we compute the features for both the source and target domains, and then we train K-Means on the source domain, with the number of clusters equal to the number of classes. We take the centroids which result after applying K-Means on the source domain and use them as initialization for the target domain. After training, we generate the pseudo-labels used further during each mini-batch step.
For the CDCL method, the experimental setup is similar to the one used for the CAT. We varied the temperature $\tau \in \{ 0.1, 0.5, 1 \}$ and the coefficient $\gamma \in \{ 0, 0.1, 1, 5 \}$. 
Table~\ref{tab:cdcl_results} provides the results of our analysis. We observe that both $\tau$ and $\gamma$ affect the performance. The best results were attained when $\tau=1$, and $\gamma=5$, achieving 63.9\% accuracy on the target domain, while $\tau=0.5$ generates the best values on the source dataset. It proves that $L_{CDC}$ performs some regularization on the source domain.
We noticed that the models often produce a high false positive rate, affecting the recall more than the precision.
In addition, training for more epochs, the model starts overfitting on both source and target domains while degrading the performance of the validation set.

\begin{table}[!t]
\centering
\caption{\label{tab:cdcl_results} Results for the CDCL framework on BuzzFeed (source) to Hyperpartisan-L (target) adaptation.}
\resizebox{0.8\columnwidth}{!}{
\begin{tabular}{l|l|c|c|c|c}
\hline
\multicolumn{1}{c|}{\multirow{2}{*}{\textbf{$\tau$}}} & \multicolumn{1}{c|}{\multirow{2}{*}{\textbf{$\gamma$}}} & \multicolumn{2}{c|}{\textbf{Source}} & \multicolumn{2}{c}{\textbf{Target}} \\ \cline{3-6}
\multicolumn{1}{c|}{} & \multicolumn{1}{c|}{} & \multicolumn{1}{c|}{\textbf{Acc(\%)}} & \multicolumn{1}{c|}{\textbf{F1(\%)}} & \multicolumn{1}{c|}{\textbf{Acc(\%)}} & \multicolumn{1}{c}{\textbf{F1(\%)}} \\ \hline

0.1 & 0   & 95.6          & 95.2          & 59.9          & 62.2 \\
0.1 & 0.1 & 91.3          & 90.1          & 63.3          & 64.9 \\
0.1 & 1   & 96.2          & 96.0          & 61.9          & 68.8 \\
0.1 & 5   & 96.2          & 96.0          & 62.6          & 67.9 \\
\hline
0.5 & 0   & 95.0          & 95.2          & 60.4          & 64.3 \\
0.5 & 0.1 & 95.3          & 95.7          & 57.1          & 67.8 \\
0.5 & 1   & 89.4          & 89.6          & 60.8          & 63.9 \\
0.5 & 5   & \textbf{96.5} & \textbf{96.4} & 63.4          & 66.5 \\
\hline
1   & 0   & 95.9          & 95.8          & 63.3          & 65.2 \\
1   & 0.1 & 95.9          & 95.8          & 61.6          & 68.6 \\
1   & 1   & 92.2          & 92.6          & 61.9          & 67.3 \\
1   & 5   & 95.6          & 95.4          & \textbf{63.9} & \textbf{69.2} \\
\hline
\end{tabular}}
\end{table}

\subsection{Results for Text Augmentation Based on TF-IDF}

We explore a data augmentation technique based on TF-IDF as proposed by~\citet{oord2019representation} for consistency training. 
Thus, we compute the TF-IDF score for every token from the corpus and associate it with the probability of it being changed. %– the higher the probability, the higher the chance of being selected
The words with the higher probability are replaced with non-keywords from the vocabulary to avoid changing the meaning of the text. The TF-IDF-based word replacement depends on a hyperparameter $p$ that controls the level of augmentation enabled on the dataset.
We vary $p$ for our experiments to augment the BuzzFeed dataset with multiple augmentation levels.
%See Table \ref{tab:aug_results} for the configurations and results for all three training approaches. Where we indicate two or three values per augmentation type, we applied each value of $p$ and concatenated the augmented examples over the original dataset. 
Table \ref{tab:aug_results} shows the results for all training configurations, where two or three values per augmentation type indicate that we applied each value of $p$ and concatenated the augmented examples over the original dataset. 
Also, zero suggests that only the unaltered dataset was used. Using more augmentations (e.g., $p \in \{0.1, 0.2, 0.3\}$) on the CDCL and CAT frameworks yields better overall results, while on UDA, using a much stronger augmentation (i.e., $p=0.5$) leads to better results.

One problem with this data augmentation technique is that it may alter the text in a way that is not coherent anymore, specifically when many tokens are changed. The most frequent words may not always have the same meaning, so their contextualized representation is affected.  Since the context defines the meaning of a word in language models, this augmentation changes the representation, especially on unlabelled data. Table~\ref{tab:aug_results} illustrates the issue on the target dataset. However, on the source dataset, the performance is not affected but generally improved. %The less affected approach seems to be CAT, which might be because we employ pseudo-labels that are updated over time for the target dataset.

\begin{table}[!t]
\centering
\caption{\label{tab:aug_results} Results for the TF-IDF-based data augmentation. The source is BuzzFeed and the target is Hyperpartisan-L.}
\resizebox{0.8\columnwidth}{!}{
\begin{tabular}{l|c|c|c|c}
\hline
\multicolumn{1}{c|}{\multirow{2}{*}{\textbf{$p$}}} & \multicolumn{2}{c|}{\textbf{Source}} & \multicolumn{2}{c}{\textbf{Target}} \\ \cline{2-5}
\multicolumn{1}{c|}{} & \multicolumn{1}{c|}{\textbf{Acc(\%)}} & \multicolumn{1}{c|}{\textbf{F1(\%)}} & \multicolumn{1}{c|}{\textbf{Acc(\%)}} & \multicolumn{1}{c}{\textbf{F1(\%)}} \\
\hline
\multicolumn{5}{c}{\textbf{UDA}} \\
\hline
   0           & 94.0  & 93.4 & 59.1  & \textbf{64.5} \\
   0.5         & 95.8  & 95.5 & \textbf{63.2}  & 62.7 \\
   0.1/0.2     & 94.7   & 94.5 & 57.3   & 61.5 \\
   0.1/0.2/0.3 & \textbf{98.4}  & \textbf{98.3} & 61.3  & 46.9 \\
\hline
\multicolumn{5}{c}{\textbf{CAT}} \\
\hline
   0            & 95.9  & 95.7 & 59.9   & 61.5 \\
   0.5          & 93.0  & 92.7 & 60.5   & \textbf{65.2} \\
   0.1/0.2      & \textbf{98.8}  & \textbf{98.8} & \textbf{62.7}  & 64.0 \\
   0.1/0.2/0.3  & 98.2  & 98.1 & 60.7  & 64.7 \\
\hline
\multicolumn{5}{c}{\textbf{CDCL}} \\
\hline
   0            & 94.0  & 93.4 & 60.8  & 69.4 \\
   0.5          & 95.1  & 94.8 & 63.2  & 69.0 \\
   0.1/0.2      & 97.3  & 97.3  & 63.6  & 68.9 \\
   0.1/0.2/0.3 & \textbf{98.8} & \textbf{98.8} & \textbf{64.4}  & \textbf{69.4} \\
 \hline
\end{tabular}}
\end{table}

\begin{table*}[!ht]
\caption{\label{tab:cb_uda} Results for the cluster- and topic-based UDA, where 0, 1, and 2 identify cluster/topic assignments given by the algorithm. The best score for each line is underlined, while bold indicates the best overall metrics.}
\centering
\resizebox{\textwidth}{!}{
\begin{tabular}{l|c|c|c|c|c|c|c|c|c|c|c|c}
\hline
\multicolumn{1}{c|}{\multirow{2}{*}{\textbf{Method}}} & \multicolumn{2}{c|}{\textbf{0 $\rightarrow$ 1}}                      & \multicolumn{2}{c|}{\textbf{1 $\rightarrow$ 0}}                      & \multicolumn{2}{c|}{\textbf{2 $\rightarrow$ 0}}                      & \multicolumn{2}{c|}{\textbf{0 $\rightarrow$ 2}}                             & \multicolumn{2}{c|}{\textbf{1 $\rightarrow$ 2}}                      & \multicolumn{2}{c}{\textbf{2 $\rightarrow$ 1}}                      \\ \cline{2-13} 
\multicolumn{1}{c|}{}                                 & \multicolumn{1}{c|}{\textbf{Acc(\%)}} & \multicolumn{1}{c|}{\textbf{F1(\%)}} & \multicolumn{1}{c|}{\textbf{Acc(\%)}} & \multicolumn{1}{c|}{\textbf{F1(\%)}} & \multicolumn{1}{c|}{\textbf{Acc(\%)}} & \multicolumn{1}{c|}{\textbf{F1(\%)}} & \multicolumn{1}{c|}{\textbf{Acc(\%)}}        & \multicolumn{1}{c|}{\textbf{F1(\%)}} & \multicolumn{1}{c|}{\textbf{Acc(\%)}} & \multicolumn{1}{c|}{\textbf{F1(\%)}} & \multicolumn{1}{c|}{\textbf{Acc(\%)}} & \multicolumn{1}{c}{\textbf{F1(\%)}} \\ \hline
K-Means-euclidean                                   & {67.2}         & 68.2                 & {66.1}         & 68.6            & {64.1}         & 65.6                   & {\ul \textbf{67.9}}   & {\ul 69.3}         & {61.9}         & 68.0          & {65.4}         & 69.2        \\
K-Means-cosine                                 & {64.2}         & {69.0}               & {63.5}         & {\ul 70.0}      & {66.0}         & {63.6}                 & {\ul 66.3}            & {67.8}             & {64.1}         & 68.5          & {62.4}         & {67.3}      \\
K-Medoids                                    & {66.0}         & 64.2                 & {62.7}         & 57.8            & {\ul 66.3}     & {\ul 68.0}             & {64.2}                & 57.5               & {61.7}         & 52.1          & {63.5}         & 60.9        \\ 
Gaussian Mixture                             & {\ul 67.1}     & {\ul \textbf{70.6}}  & {59.5}         & 67.7            & {{57.9}}       & {64.0}                 & {64.9}                & 69.6               & {59.7}         & 68.0          & {65.3}         & 64.2        \\
HDBSCAN                                     & {\ul 65.1}     & {\ul 68.9}           & {62.5}         & 63.4            & {50}           & {0.0}                  & {60.0}                & 55.6               & {62.2}         & 66.0          & {50.0}         & 0.0         \\ 
\hline
LDA                                         & {61.8}         & 52.2                 & {59.0}         & 43.5            & {\ul 66.1}     & 61.9                   & {62.6}                & {\ul 66.2}         & {49.4}         & 61.9          & {59.8}         & 46.2        \\
NMF                                         & {\ul 63.3}     & 53.3                 & {59.9}         & 55.7            & {56.0}         & {\ul 58.1}             & {54.9}                & 36.3               & {59.8}         & 57.0          & {60.5}         & 45.4        \\
LSA                                         & {\ul 62.1}     & {\ul 70.3}           & {50.0}         & 66.4            & {51.5}         & 8.6                    & {51.6}                & 65.6               & {53.1}         & 64.6          & {61.4}         & 70.0        \\ 
pLSA                                         & {61.6}         & {\ul 68.7}           & {50.0}         & 1.4             & {57.1}         & 66.1                   & {60.1}                & 66.2               & {60.2}         & 54.8          & {\ul 62.4}     &  67.6       \\
\hline
\end{tabular}}
\end{table*}

\subsection{Results for Cluster- and Topic-Based UDA}
\label{sec:cb_uda}

In the topic-based UDA approach, we follow the same experimental setup as in classical UDA. For training, the only difference is that we train all models for 10 epochs. We explore both, the clustering on RoBERTa features (i.e., K-Means with Euclidean or cosine distance, K-Medoids, Gaussian Mixture, and HDBSCAN) and the topic modeling algorithms on TF-IDF features (i.e., LDA, NMF, LSA, and pLSA) to split the representation. 
%To obtain better clustering and avoid bad optima, we use the K-Means++~\cite{kmeans++} initialization.
We evaluate the experiments on the Hyperpartisan-L test set and present the results in Table~\ref{tab:cb_uda}. Using clustering algorithms for domain labels provides the best overall results compared to Table~\ref{tab:UDA_results}. The best-performing models outperform the UDA approach by over 3\% in accuracy and are obtained when we adapted from a larger to a smaller split. It is noteworthy that for the HDBSCAN, the cluster 2 contains very few annotated examples (i.e., 332) compared with the other two (i.e., 17,092 and 12,576), resulting in adaptation failure. When using the topic modeling, we see a degradation in performance, especially in the case of NMF. 
Compared with the RoBERTa baseline (see Table~\ref{tab:baseline-results-2}), the model achieves similar F1-scores.
% We think this approach achieves better results because we utilize some of the labels from the Hyperpartisan-dataset dataset, leading to similar distributions for domain shift minimization.

% \begin{table}[!h]
% \caption{\label{tab:cb_uda} Results of UDA considering K-Means clusters as domains}
% \centering
% \begin{tabular}{|c|c|c|c|c|}
% \hline
% \multicolumn{1}{|l|}{\textbf{Adaptation}} &
%   \multicolumn{1}{l|}{\textbf{Acc(\%)}} &
%   \multicolumn{1}{l|}{\textbf{P(\%)}} &
%   \multicolumn{1}{l|}{\textbf{R(\%)}} &
%   \multicolumn{1}{l|}{\textbf{F1(\%)}} \\ \hline
% 0 $\rightarrow$ 1 & 67.2          & 66.2          & 70.5          & 68.2          \\
% 1 $\rightarrow$ 0 & 66.1          & 63.9          & 74.0          & 68.6          \\
% 2 $\rightarrow$ 0 & 64.1          & 63.0          & 68.3          & 65.6          \\
% 0 $\rightarrow$ 2 & \textbf{67.9} & \textbf{66.4} & 72.5          & \textbf{69.3} \\
% 1 $\rightarrow$ 2 & 61.9          & 58.5          & \textbf{81.2} & 68.0          \\
% 2 $\rightarrow$ 1 & 65.4          & 62.3          & 77.9          & 69.2          \\ \hline
% \end{tabular}
% \end{table}

% Unsupervised domain adaptation performed on each combination of groups of points generated using K-Means, on the Hyperpartisan-L test set. The indices identify the clusters. With bold we indicate the highest scores at each metric.

\subsection{Feature Visualization}
\label{sec:feature_vis}

\begin{figure}[!t]
\centering
\subfloat[Baseline]{\includegraphics[width=0.45\columnwidth]{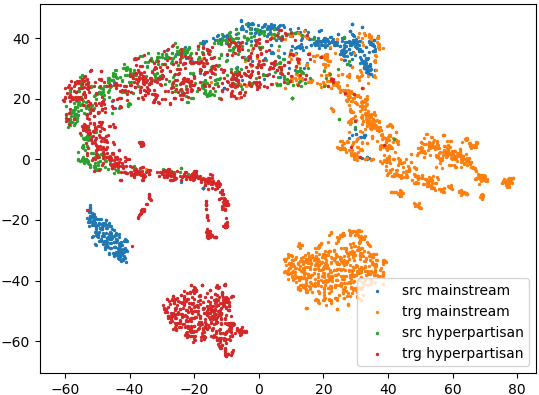}}
\hfil
\subfloat[UDA]{\includegraphics[width=0.45\columnwidth]{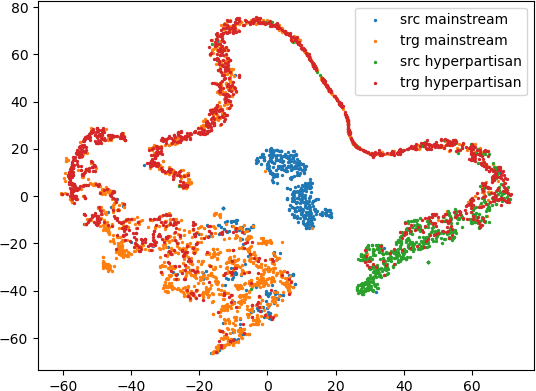}}
\hfil
\subfloat[CAT]{\includegraphics[width=0.45\columnwidth]{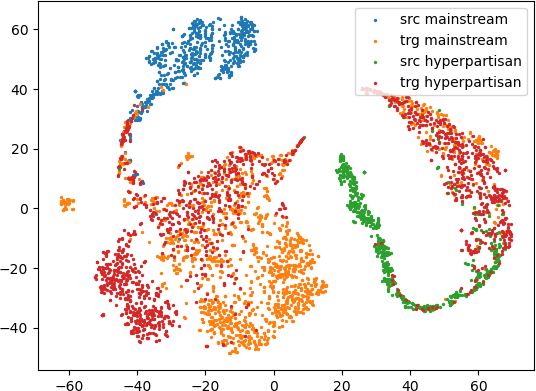}}
\hfil
\subfloat[CDCL]{\includegraphics[width=0.45\columnwidth]{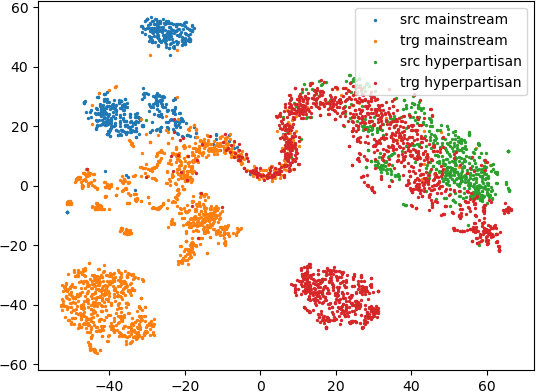}}

\caption{\label{fig:tsne} t-SNE visualizations of the feature representations for the BuzzFeed (source) and Hyperpartisan-L (target) datasets. 
Blue -- source (src) mainstream, orange -- target (trg) mainstreams, green -- source hyperpartisan, and red -- target hyperpartisan. Best viewed in color.
}
\end{figure}

\begin{figure}[!t]
\centering
\subfloat[K-Means]{\includegraphics[width=0.45\columnwidth]{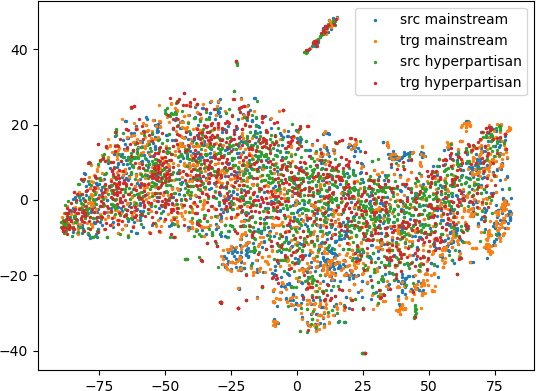}}
\hfil
\subfloat[K-Medoids]{\includegraphics[width=0.45\columnwidth]{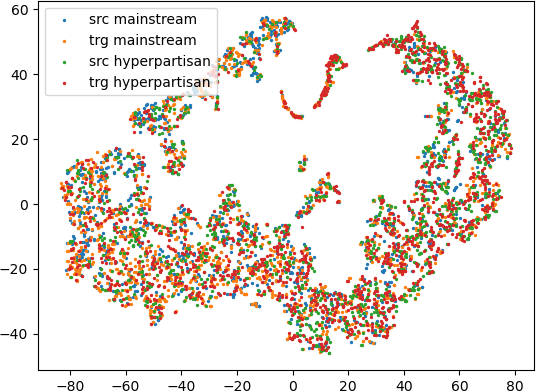}}
\hfil
\subfloat[LDA]{\includegraphics[width=0.45\columnwidth]{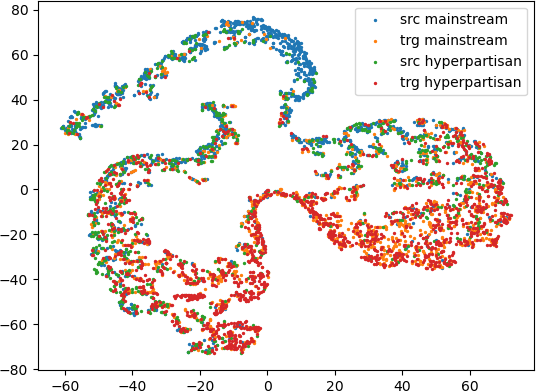}}
\hfil
\subfloat[NMF]{\includegraphics[width=0.45\columnwidth]{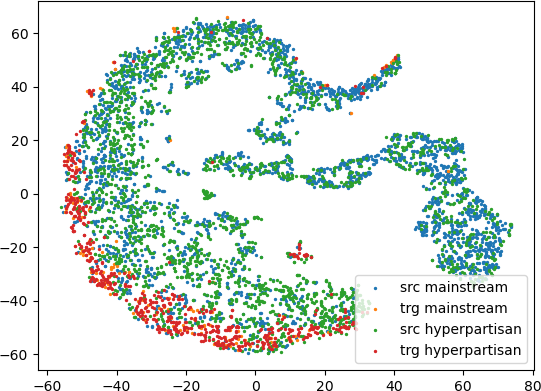}}

\caption{\label{fig:tsne_2} t-SNE visualizations of the feature representations when employing topic/clustering methods on the validation sets. Blue -- source (src) mainstream, orange -- target (trg) mainstreams, green -- source hyperpartisan, and red -- target hyperpartisan. Best viewed in color.
}
\end{figure}

We use t-SNE~\cite{vanDerMaaten2008} to visualize the feature representation learned by the best models we obtained for each category. In Figure~\ref{fig:tsne}, we present the plots for the baseline, the UDA, the CAT, and the CDCL. Using different approaches to domain adaptation may reduce the domain gap in the feature space between the two domains. Still, many examples cluster together far apart from their counterparts. UDA obtains better representations than the other methods. When considering the topic-based adaptation (see Figure~\ref{fig:tsne_2}), we notice a better separation when employing topic models. Also, we achieve poor separation among classes for K-Means and K-Medoids.

\section{Conclusions}
\label{sec:conclusions}

In this work, we addressed the problem of transferring knowledge from fake to hyperpartisan news detection. We employed three types of architectures based on unsupervised training. We conducted multiple experiments, showing the effects of the hyperparameters in the given configuration. All employed methods manage to perform some domain adaptation. In particular, we showed that CDCL obtains the best results after applying data augmentation based on TF-IDF word replacement. In contrast, CAT managed the poorest results. By analyzing the t-SNE visualization, this model did not learn a good feature representation, with a minimal domain gap between the source and target datasets. The low accuracy we hypothesize is due to a lack of data from the source domain, as we have seen that data augmentation helped. For future work, we aim to investigate our approaches on other fake news datasets. %Nonetheless, domain adaptation is an effective technique for improving the performances of baseline models.

\section*{Acknowledgments}
This research has been funded by the University Politehnica of Bucharest through the PubArt program.

\bibliographystyle{acl_natbib}
\bibliography{anthology,ranlp2023}

\clearpage
\appendix

\begin{table*}[!th]
\caption{\label{tab:gpt2_aug} Results for the text augmentation using GPT-2. The source is BuzzFeed and the target is Hyperpartisan-L.}
\centering
\resizebox{\textwidth}{!}{
\begin{tabular}{l|l|cccc|cccc|cccc}
\hline
\multicolumn{1}{c|}{\multirow{3}{*}{\textbf{\begin{tabular}[c]{@{}c@{}}Decoding\\ Strategy\end{tabular}}}} & \multicolumn{1}{c|}{\multirow{3}{*}{$T$}} & \multicolumn{4}{c|}{\textbf{UDA}}                                                                                                      & \multicolumn{4}{c|}{\textbf{CAT}}                                                                                                      & \multicolumn{4}{c}{\textbf{CDCL}}                                                                                                      \\ \cline{3-14} 
\multicolumn{1}{c|}{}                                                                                      & \multicolumn{1}{c|}{}                                    & \multicolumn{2}{c|}{\textbf{Source}}                                         & \multicolumn{2}{c|}{\textbf{Target}}                    & \multicolumn{2}{c|}{\textbf{Source}}                                         & \multicolumn{2}{c|}{\textbf{Target}}                    & \multicolumn{2}{c|}{\textbf{Source}}                                         & \multicolumn{2}{c}{\textbf{Target}}                     \\ \cline{3-14} 
\multicolumn{1}{c|}{}                                                                                      & \multicolumn{1}{c|}{}                                    & \multicolumn{1}{c|}{\textbf{Acc(\%)}} & \multicolumn{1}{c|}{\textbf{F1(\%)}} & \multicolumn{1}{c|}{\textbf{Acc(\%)}} & \textbf{F1(\%)} & \multicolumn{1}{c|}{\textbf{Acc(\%)}} & \multicolumn{1}{c|}{\textbf{F1(\%)}} & \multicolumn{1}{c|}{\textbf{Acc(\%)}} & \textbf{F1(\%)} & \multicolumn{1}{c|}{\textbf{Acc(\%)}} & \multicolumn{1}{c|}{\textbf{F1(\%)}} & \multicolumn{1}{c|}{\textbf{Acc(\%)}} & \textbf{F1(\%)} \\ \hline
\multirow{3}{*}{Greedy decoding}                                                                                    & 3                                                        & \multicolumn{1}{c|}{96.0}             & \multicolumn{1}{c|}{95.6}            & \multicolumn{1}{c|}{62.5}             & 68.3            & \multicolumn{1}{c|}{96.6}             & \multicolumn{1}{c|}{95.9}            & \multicolumn{1}{c|}{63.7}             & 65.9            & \multicolumn{1}{c|}{97.2}             & \multicolumn{1}{c|}{97.2}            & \multicolumn{1}{c|}{\textbf{64.4}}    & \textbf{70.4}   \\
                                                                                                           & 3/5                                                      & \multicolumn{1}{c|}{96.0}             & \multicolumn{1}{c|}{95.6}            & \multicolumn{1}{c|}{60.1}             & \textbf{69.2}   & \multicolumn{1}{c|}{96.6}             & \multicolumn{1}{c|}{95.9}            & \multicolumn{1}{c|}{63.9}             & 66.6            & \multicolumn{1}{c|}{96.3}             & \multicolumn{1}{c|}{96.3}            & \multicolumn{1}{c|}{61.7}             & 68.6            \\
                                                                                                           & 3/5/10                                                   & \multicolumn{1}{c|}{96.3}             & \multicolumn{1}{c|}{96.1}            & \multicolumn{1}{c|}{55.4}             & 67.3            & \multicolumn{1}{c|}{95.0}             & \multicolumn{1}{c|}{94.1}            & \multicolumn{1}{c|}{63.2}             & 66.5            & \multicolumn{1}{c|}{97.2}             & \multicolumn{1}{c|}{97.2}            & \multicolumn{1}{c|}{62.6}             & 68.8            \\ \hline
\multirow{3}{*}{Beam search}                                                                               & 3                                                        & \multicolumn{1}{c|}{95.7}             & \multicolumn{1}{c|}{95.3}            & \multicolumn{1}{c|}{\textbf{63.4}}    & 68.2            & \multicolumn{1}{c|}{94.4}             & \multicolumn{1}{c|}{93.1}            & \multicolumn{1}{c|}{63.5}             & 64.1            & \multicolumn{1}{c|}{94.7}             & \multicolumn{1}{c|}{94.5}            & \multicolumn{1}{c|}{64.2}             & 68.1            \\
                                                                                                           & 3/5                                                      & \multicolumn{1}{c|}{95.7}             & \multicolumn{1}{c|}{95.3}            & \multicolumn{1}{c|}{57.1}             & 68.4            & \multicolumn{1}{c|}{94.4}             & \multicolumn{1}{c|}{93.1}            & \multicolumn{1}{c|}{64.2}             & 63.4            & \multicolumn{1}{c|}{96.6}             & \multicolumn{1}{c|}{96.6}            & \multicolumn{1}{c|}{61.5}             & 68.0            \\
                                                                                                           & 3/5/10                                                   & \multicolumn{1}{c|}{\textbf{97.8}}    & \multicolumn{1}{c|}{\textbf{97.7}}   & \multicolumn{1}{c|}{62.1}             & 68.9            & \multicolumn{1}{c|}{96.3}             & \multicolumn{1}{c|}{95.6}            & \multicolumn{1}{c|}{\textbf{64.4}}    & 66.1            & \multicolumn{1}{c|}{96.9}             & \multicolumn{1}{c|}{96.9}            & \multicolumn{1}{c|}{60.7}             & 66.2            \\ \hline
\multirow{3}{*}{Top-k}                                                                                     & 3                                                        & \multicolumn{1}{c|}{94.7}             & \multicolumn{1}{c|}{94.2}            & \multicolumn{1}{c|}{62.9}             & 65.4            & \multicolumn{1}{c|}{96.3}             & \multicolumn{1}{c|}{96.2}            & \multicolumn{1}{c|}{62.6}             & 66.7            & \multicolumn{1}{c|}{96.0}             & \multicolumn{1}{c|}{95.9}            & \multicolumn{1}{c|}{61.6}             & 68.3            \\
                                                                                                           & 3/5                                                      & \multicolumn{1}{c|}{95.0}             & \multicolumn{1}{c|}{94.7}            & \multicolumn{1}{c|}{61.7}             & 68.6            & \multicolumn{1}{c|}{96.9}             & \multicolumn{1}{c|}{96.8}            & \multicolumn{1}{c|}{63.8}             & 65.9            & \multicolumn{1}{c|}{96.9}             & \multicolumn{1}{c|}{96.8}            & \multicolumn{1}{c|}{60.7}             & 66.7            \\
                                                                                                           & 3/5/10                                                   & \multicolumn{1}{c|}{96.3}             & \multicolumn{1}{c|}{96.0}            & \multicolumn{1}{c|}{60.0}             & 68.5            & \multicolumn{1}{c|}{\textbf{97.2}}    & \multicolumn{1}{c|}{\textbf{97.2}}   & \multicolumn{1}{c|}{63.6}             & \textbf{68.4}   & \multicolumn{1}{c|}{96.6}             & \multicolumn{1}{c|}{96.5}            & \multicolumn{1}{c|}{61.7}             & 69.0            \\ \hline
\multirow{3}{*}{Top-p}                                                                                     & 3                                                        & \multicolumn{1}{c|}{95.7}             & \multicolumn{1}{c|}{95.3}            & \multicolumn{1}{c|}{61.5}             & 67.1            & \multicolumn{1}{c|}{96.9}             & \multicolumn{1}{c|}{96.8}            & \multicolumn{1}{c|}{63.3}             & 61.4            & \multicolumn{1}{c|}{97.2}             & \multicolumn{1}{c|}{97.1}            & \multicolumn{1}{c|}{63.6}             & 69.1            \\
                                                                                                           & 3/5                                                      & \multicolumn{1}{c|}{95.0}             & \multicolumn{1}{c|}{94.7}            & \multicolumn{1}{c|}{62.1}             & 67.1            & \multicolumn{1}{c|}{96.3}             & \multicolumn{1}{c|}{96.1}            & \multicolumn{1}{c|}{62.6}             & 62.9            & \multicolumn{1}{c|}{\textbf{97.8}}    & \multicolumn{1}{c|}{\textbf{97.7}}   & \multicolumn{1}{c|}{62.3}             & 68.1            \\
                                                                                                           & 3/5/10                                                   & \multicolumn{1}{c|}{95.7}             & \multicolumn{1}{c|}{95.3}            & \multicolumn{1}{c|}{61.4}             & 68.3            & \multicolumn{1}{c|}{96.3}             & \multicolumn{1}{c|}{96.2}            & \multicolumn{1}{c|}{61.5}             & 59.3            & \multicolumn{1}{c|}{97.5}             & \multicolumn{1}{c|}{97.5}            & \multicolumn{1}{c|}{62.5}             & 67.9            \\ \hline
\end{tabular}}
\end{table*}

\section{Appendix}

\subsection{Results for Text Augmentation Based on GPT-2}
\label{app:gpt2_results}

Observing the improvements obtained using TF-IDF augmentation, we consider text generation an alternative. Therefore, we employ the GPT-2 model~\cite{radford2019language} to conditionally generate new examples given the news types (i.e., left-wing, right-wing, and mainstream). We follow an approach similar to the LAMBADA method proposed by \citet{lambada}. Therefore, we fine-tune the GPT-2 base model on the hyperpartisan Buzzfeed dataset to generate new samples. Inspired by other works \cite{NEURIPS2020_1457c0d6,liu2023pre,niculescu2022summary}, we build the pre-training dataset using, for each sample, the following prompt:
\begin{align*}
&\texttt{News type : <LABEL>} \\
&\texttt{Text : <TEXT>} \\
&\texttt{<|endoftext|>}
\end{align*}
where \texttt{<LABEL>} is \textit{left}, \textit{right}, or \textit{mainstream}, \texttt{<TEXT>} is the news content, and \texttt{<|endoftext|>} is the end token of the text.
Since we use a relatively small context during experiments (i.e., 128 tokens), we do not require the auto-regressive model to learn to generate long samples, but rather more variation within the generated samples. To achieve this, we split each text into sentences and group every three sentences into one example under the same label.

As suggested by \citet{kumar2020data}, during data generation, we iterate over each sample from the training set and prompt the model with \texttt{News type: <LABEL> Text:} followed by the first $T$ tokens from each sample. Because the model may generate text that is not correlated with the label (i.e., either the model ignores the prompt label~\cite{webson-pavlick-2022-prompt}, or there is not enough data for the model to learn a clear distinction), we use the RoBERTa baseline model fine-tuned on the Buzzfeed dataset to filter the samples, ignoring those that do not match the model's prediction.

Text generation quality depends on the decoding strategy; thus, we explore multiple approaches. 

\textbf{Greedy decoding.} The most trivial and fastest way of synthesizing text is to consider the token with the highest probability. Albeit simple, it has the disadvantage of generating repetitive and missing higher probability words behind lower probability ones.

\textbf{Beam search.} Beam search~\cite{freitag-al-onaizan-2017-beam} seeks to solve the low probability issue from the greedy decoding by choosing the highest probability sequence within a number of beams. This method generally yields to higher probability sequence than greedy decoding. During experiments, we set the number of beams to 5.

\textbf{Top-k.} Using the top-k decoding~\cite{fan-etal-2018-hierarchical}, we consider only the highest $k$ next tokens from the probability distribution over possible next tokens. This simple yet effective method produces more human-like text than previous approaches. In our experiments, we consider $k=30$ tokens.

\textbf{Top-p nucleus sampling.} Introduced by \citet{HoltzmanBDFC20}, the top-p nucleus sampling is an extension over top-k. We choose the tokens from the smallest subset whose cumulative probability is at least $p$ instead of choosing from the top $k$ probabilities. For experiments, we set $p=99\%$.

To generate more samples, we repeat the procedure while setting $T \in \{3,5,10\}$. The results are shown in Table~\ref{tab:gpt2_aug}. CDCL obtains the highest scores on the source and target datasets using top-p and greedy decoding, respectively. On the source dataset, the accuracy reaches 97.8\% and the F1-score tops at 97.7\%, while on the target dataset, the best accuracy is 64.4\% and F1-score is 70.4\%. Compared with the TF-IDF text augmentation, the GPT-2 augmentation produces a higher best F1-score by 1\% on the target test set, and achieves lower scores on the source test set by 1\%. 
In addition, we notice that the performance improves when adding more data, especially on the source dataset, where we see an average improvement of 0.6\% and 0.8\% for accuracy and F1-score, respectively. On average, greedy decoding improves the target F1-score (i.e., 68.0$\pm$1.5\%) while the lowest average is obtained by top-p (i.e., 65.7$\pm$3.5\%). We notice a small improvement in favor of top-p compared with top-k on the source domain, but the target domain does not benefit from it in our case.

\end{document}